\documentclass[conference]{IEEEtran}
\IEEEoverridecommandlockouts
\usepackage{cite}
\usepackage{amsmath,amssymb,amsfonts}
\usepackage{algorithmic}
\usepackage{graphicx}
\usepackage{textcomp}
\usepackage{makecell}
\usepackage{xcolor}
\usepackage{multirow}
\usepackage{url}
\def\BibTeX{{\rm B\kern-.05em{\sc i\kern-.025em b}\kern-.08em
    T\kern-.1667em\lower.7ex\hbox{E}\kern-.125emX}}
\begin{document}

\title{Enhancing Authorship Attribution with Synthetic Paintings

\author{
\IEEEauthorblockN{
    Clarissa Loures\IEEEauthorrefmark{1}\IEEEauthorrefmark{2},
    Caio Hosken\IEEEauthorrefmark{1}\IEEEauthorrefmark{2},
    Luan Oliveira\IEEEauthorrefmark{1}\IEEEauthorrefmark{2},
    Gianlucca Zuin\IEEEauthorrefmark{1}\IEEEauthorrefmark{2}
    Adriano Veloso\IEEEauthorrefmark{1}\IEEEauthorrefmark{2}
}
\IEEEauthorblockA{\IEEEauthorrefmark{1}Universidade Federal de Minas Gerais (UFMG), Brazil}
\IEEEauthorblockA{\IEEEauthorrefmark{2}Instituto Kunumi, Brazil}
\IEEEauthorblockA{Email: \{clarissa.loures, caio.hosken, luan.oliveira, gianlucca, adriano\}@kunumi.com}
}

\thanks{This work was funded by the authors' individual grants from Kunumi.}
}

\maketitle

\begin{abstract}
Attributing authorship to paintings is a historically complex task, and one of its main challenges is the limited availability of real artworks for training computational models. This study investigates whether synthetic images, generated through DreamBooth fine-tuning of Stable Diffusion, can improve the performance of classification models in this context. We propose a hybrid approach that combines real and synthetic data to enhance model accuracy and generalization across similar artistic styles. Experimental results show that adding synthetic images leads to higher ROC-AUC and accuracy compared to using only real paintings. By integrating generative and discriminative methods, this work contributes to the development of computer vision techniques for artwork authentication in data-scarce scenarios.
\end{abstract}


\section{Introduction}

Authenticity is a fragile and elusive concept, yet it is crucial for both the cultural and financial value of an artwork \cite{b1}. Traditional authentication methods rely on experts with extensive knowledge of art history, who analyze an artist's life and work to determine the authenticity of a piece \cite{b2}. In addition to expert evaluation, physical techniques such as painting sampling \cite{b7} and X-ray radiography \cite{b6} are commonly used to uncover underlying details that may indicate originality or forgery. More recently, mathematical \cite{b2} and computational \cite{b3, b4} methods have emerged as complementary approaches, offering new perspectives and tools to enhance the accuracy and efficiency of artwork authentication. Traditional CNN-based approaches have shown strong performance in painting authentication but remain limited by their dependence on large, diverse datasets \cite{b9, b10}.

To overcome the limitation of small datasets in image classification, two effective approaches have emerged: data augmentation and synthetic data generation. Data augmentation improves model robustness by applying transformations, such as rotation, scaling, and contrast adjustments, to real images, generating diverse variations of the original data \cite{b12}. In contrast, synthetic data generation uses advanced generative models, including \cite{b13} and diffusion models \cite{b14}, to create new images that emulate the artistic style of real paintings. These models learn stylistic patterns and visual structures from input data, producing images that aim to maintain the distinctive characteristics of the original artworks.

Diffusion models offer several advantages, particularly in stability and controllability \cite{b18}. Unlike GANs, which generate images in a single pass, diffusion models follow a stepwise denoising process, progressively refining an image from noise. This reduces training instability and allows for finer control over the generated outputs. 

The representation of images in a latent space using visual embeddings plays a central role in our classification pipeline. We employ three state-of-the-art deep learning architectures to extract these embeddings: MaxViT \cite{b17}, which combines convolutional layers and transformer mechanisms to capture spatial and hierarchical features; BEiT v2  \cite{b16}, based on masked image modeling for self-supervised learning; and VOLO \cite{b15}, which enhances spatial attention through the Vision Outlooker structure. These embeddings provide compact and discriminative representations that serve as input for training lightweight yet accurate classification models.

This paper investigates the use of synthetic images generated by diffusion models to enhance binary classification LightGBM models for authorship verification. To achieve this, we employ a small dataset, ranging from 7 to 25 images per artist, focusing on seven British artists who lived during the same historical period and region: Gainsborough Dupont (GD), George Romney (GR), Thomas Gainsborough (TG), George Morland (GM), James Northcote (JN), Thomas Barker (TB) and John Hoppner (JH).

These artists present a particularly challenging scenario for authorship attribution due to their shared historical and geographical context, as they were all active in Britain during the late 18th and early 19th centuries, often depicting similar subjects, using comparable materials and being influenced by overlapping artistic movements~\cite{shepherd1881short, roberts1970british, graham1999history}. This stylistic proximity makes distinguishing between them more difficult, often requiring nuanced expert analysis. Furthermore, the availability of high-quality digitized images for these artists is limited and uneven across collections, restricting the amount of training data for computational models. As a result, this group represents a compelling test case for authorship verification: the task involves a small, imbalanced dataset with visually similar but distinct artistic styles, ideal for evaluating the effectiveness of synthetic data in improving classification models.

Our goal is to demonstrate that synthetic images can support authorship attribution tasks in scenarios with limited data and subtle visual differences between classes. To this end, we fine-tune a diffusion model for each artist using a generic prompt,  ``a full \textit{owhx} painting", allowing the model to generate samples that reflect the artist’s visual style. These synthetic images are then combined with real paintings to train supervised classifiers. Through this experiment, we aim to assess how synthetic images contribute to the performance of supervised classification models in settings where visual variations are small and generalization is particularly challenging.

\section{Related Work}

Synthetic data is increasingly viewed as a practical solution for training vision models, particularly when collecting or annotating large datasets is infeasible \cite{b21}. The emergence of text-to-image diffusion models has further expanded the possibilities of dataset generation, allowing high-quality visual samples to be produced directly from prompts rather than relying on 3D renderers or simulation environments \cite{b22}.

Stable Diffusion introduced a latent-space architecture that generates high-resolution images with reduced computational cost while maintaining semantic control over the output \cite{b23}. This made it a common choice for creating stylized datasets and driving domain-specific applications. A significant advancement in personalizing these models is DreamBooth \cite{b19}, which fine-tunes a text-to-image model to learn a new concept from just a few examples. This technique associates a unique identifier with the specific subject or style, allowing for the generation of novel renditions in different contexts, making it highly suitable for capturing and replicating the distinct styles of individual artists, as explored in this work. 

Several studies have explored the use of synthetic datasets to train visual classifiers, particularly in contexts where real data is scarce or difficult to obtain. Classifiers trained on synthetic ImageNet-like data produced by diffusion models have achieved performance comparable to those trained on real datasets, highlighting the potential of prompt-driven generation for building large-scale training corpora \cite{b22}. These results suggest that, with appropriate prompt design and sufficient sample diversity, synthetic images can support robust generalization across downstream tasks.

The relationship between data quality, quantity, and model performance has also been studied in detail. While synthetic datasets generally scale less efficiently than real ones in large supervised settings, they can be particularly effective in low-data regimes or under distribution shifts \cite{b21}. To better understand this behavior, metrics for recognizability and diversity have been proposed to quantify the characteristics of synthetic samples that most influence generalization.

\section{Proposed Approach}
The proposed approach integrates diffusion-based image generation and discriminative modeling to improve authorship classification in low-data scenarios. The workflow consists of three main stages: data preparation, generative fine-tuning, and supervised classification. 

\subsection{Dataset}

Each artist’s dataset comprises between 7 and 25 digitized paintings, specifically: GD (7), GR (13), TG (23), GM (22), JN (11), TB (14), and JH (9). For the generative model training data, we extracted images crops (1024×1024) of each artist from real artworks, selected to cover stylistic elements such as backgrounds, textures, human and animal figures, and compositional structures. The goal was to capture the diversity of visual patterns without replicating full paintings or specific objects.
For evaluation, we reserved one painting per artist as the held-out test target across all experiments. The remaining works were used both to train the classification models and to fine-tune the generative model for creating synthetic paintings. This setup prevents leakage between training and testing while ensuring that the generative stage relies on the same limited real data available to the classifier.

\subsection{Sample Preparation}
To prepare the data, for the Stable Diffusion finetunin 200 mage crops (1024×1024) extracted from real paintings of that artist,resized to 512×512. For the training of the classification model, the samples were resized to 224×224. Both preprocessing steps were required due to the architecture of the respective training models.

\subsection{Fine-Tuning with DreamBooth}

To generate stylistically consistent synthetic images, a pre-trained Stable Diffusion model was fine-tunedusing the DreamBooth framework. DreamBooth enables personalized generation from a small number of examples. One model was trained for each artist's style, with a unique identifier token embedded in the prompt to guide generation.

\subsection{Synthetic Dataset Generation}

After tuning, 100 new images were generated per artist using the prompt “A full [TOKEN] painting”  to guide the generation of new images along with a consistent negative prompt discouraging cropped or incomplete figures (cropped body, partial figures, close-up, cut off ). This strategy encouraged full-scene compositions while preserving stylistic coherence. Despite this, some partial compositions persisted, likely due to data imprinting from the training set.

\begin{figure}[!b]
    \centering
    \includegraphics[width=1\linewidth]{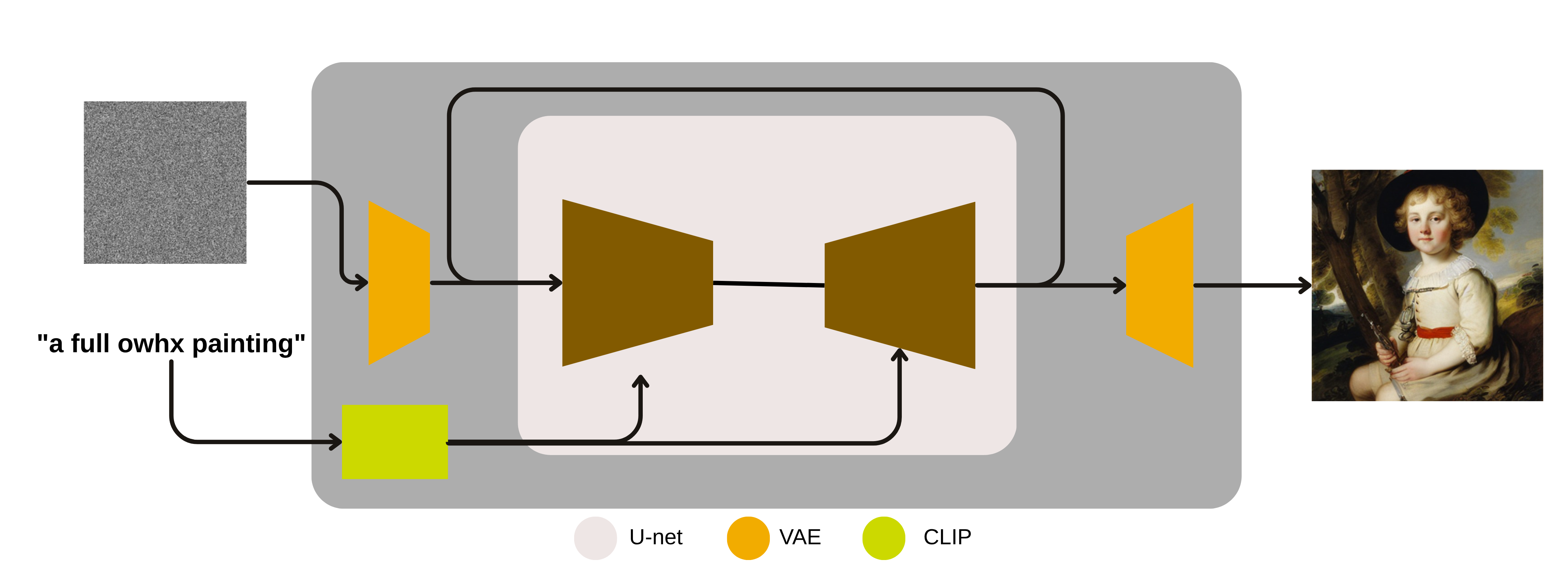}
    \caption{Overview of the image generation pipeline. Starting from Gaussian noise and a prompt (e.g., "a full owhx painting"), the CLIP encoder conditions the U-Net denoising process in latent space. After multiple refinement steps, the latent image is decoded by a VAE into a full-resolution painting. This process is repeated to generate a synthetic dataset per artist.}
    \label{fig:pipeline}
\end{figure}

As illustrated in Figure~\ref{fig:pipeline}, the generation process starts with random Gaussian noise and gradually refines it through multiple iterations using a U-Net architecture. The CLIP encoder processes the textual prompt and conditions the denoising process, guiding the image synthesis toward the desired artistic style. After iterative refinement, the latent representation is decoded into a full-resolution image via a Variational Autoencoder (VAE). This procedure is repeated for each artist to produce a diverse and stylistically coherent synthetic dataset.

\subsection{Discriminative Modeling}

To classify authorship at the patch level, we used a featurebased approach. Visual embeddings were extracted from each patch using three transformer-based models: MaxViT, BEiT v2, and VOLO. These embeddings were concatenated and used to train a binary classifier per artist using LightGBM, a gradient boosting algorithm well suited for high-dimensional tabular inputs.

Model performance was evaluated using a combination of metrics that capture both overall accuracy and class-specific behavior. The main evaluation metric was the Area Under the ROC Curve (ROC-AUC), which measures the model’s ability to distinguish between the target artist and others. In addition, we report Precision, Recall, and F1-Score for both classes, along with overall Accuracy. These metrics allow a comprehensive analysis of model behavior, including its robustness in identifying true positives (authentic patches) and true negatives (non-authentic patches), which is particularly relevant in authorship attribution tasks where class imbalance may affect performance.

\section{Experiments}

This section presents the four experimental setups designed to evaluate the impact of synthetic data on binary authorship attribution. We aim to compare different training configurations using real, synthetic, and hybrid datasets under consistent testing conditions. The key question is whether synthetic images improve model generalization, especially when real data is limited.

\subsection{Experimental Setup}

The evaluation follows a consistent protocol across all experiments: one real painting per artist is reserved for testing, and the rest are used for training. All paintings are segmented into 224×224 patches. Features are extracted using MaxViT, VOLO, and BEiT v2, and passed to a LightGBM classifier trained per artist. Models are evaluated using ROC-AUC, Accuracy, and class-specific Precision, Recall, and F1-score.

\begin{figure}
\centering
\includegraphics[width=\linewidth]{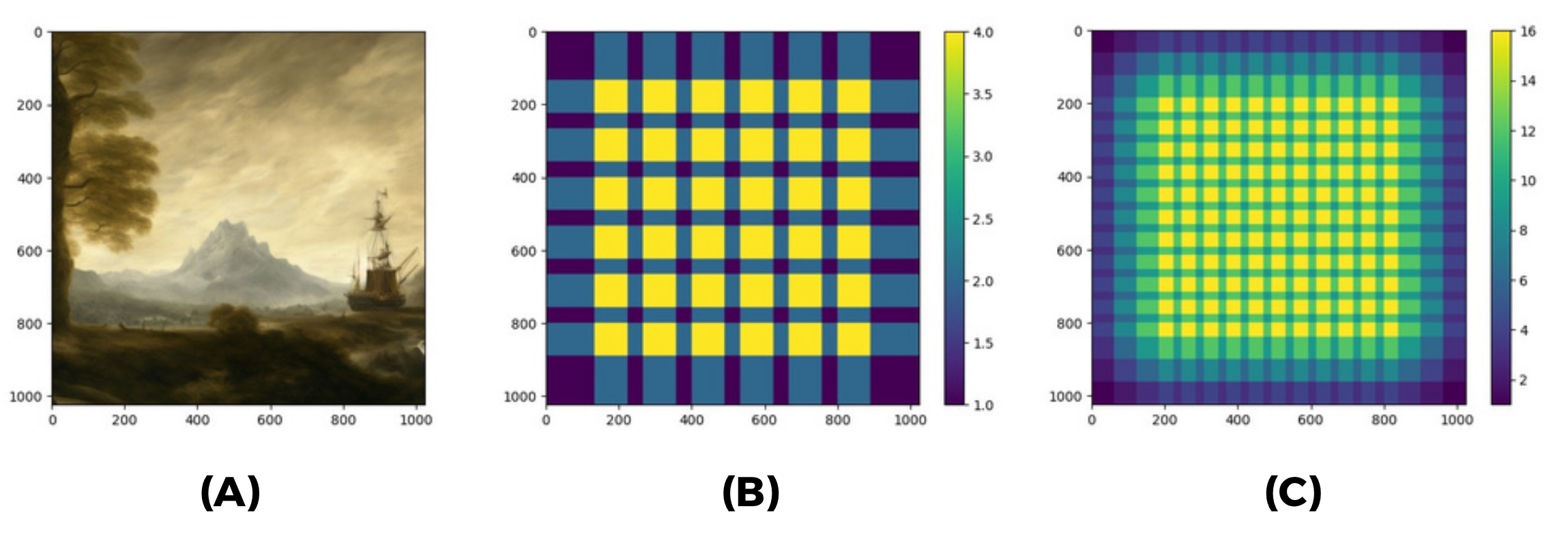}
\caption{Patch sampling strategies. (A) Original image. (B) M1: baseline sampling with moderate overlap. (C) M2: denser sampling obtained by doubling the number of patches along both axes. Axis color reflects the level of overlap between samples.}
\label{fig:patch_sampling}
\end{figure}

Patch extraction follows the \textbf{M1} sampling strategy unless otherwise specified. In M1, patches are generated with moderate overlap, and the total number of samples is calculated based on the minimum required to span the image plus an adjustment proportional to its size. In the Hybrid experiment, we also test a denser sampling variant, \textbf{M2}, which doubles the number of patches along both axes. Figure~\ref{fig:patch_sampling} illustrates the two strategies. 

We evaluate M1 and M2 to study the trade-off between coverage and redundancy. M1 provides moderate overlap and fewer patches, while M2 doubles density to capture finer stylistic cues at the cost of higher redundancy and computation. This comparison tests whether denser sampling improves generalization in data-scarce scenarios.

We propose four experiments to validate our hypothesis:

\vspace{0.1in}
\noindent\textbf{Experiment 1 (Real-Only):} This baseline experiment uses only real paintings for both training and testing. It serves as a reference to measure the benefit of synthetic augmentation. All patches are extracted using M1.

\vspace{0.1in}
\noindent\textbf{Experiment 2 (Synthetic-Only):}
In this setup, models are trained and tested exclusively on synthetic images generated via DreamBooth with Stable Diffusion. The same M1 sampling is applied to create patches. This configuration assesses the model’s upper-bound performance under perfectly aligned training and testing distributions.

\vspace{0.1in}
\noindent\textbf{Experiment 3 (Synthetic-Real):}
Here, models are trained on synthetic images but evaluated on real paintings. Like the previous experiments, patch sampling follows the M1 strategy. This setup tests the model’s ability to generalize across the domain shift introduced by synthetic data.

\vspace{0.1in}
\noindent\textbf{Experiment 4 (Real + Synthetic):}
In this final experiment, real and synthetic paintings are combined during training. The goal is to determine whether synthetic data can enhance classification when used alongside real data. We propose:
\begin{itemize}
    \item \textbf{Hybrid-M1}: a combined dataset sampled using M1.
    \item \textbf{Hybrid-M2}: a combined dataset sampled using the denser M2 strategy.
\end{itemize}

\noindent This configuration allows us to assess if increasing the number of samples through denser patch extraction improves performance when both real and synthetic images are available.

\subsection{Fine-Tuning Generative Models and Image Generation}

To ensure diversity in the training data, 200 samples were extracted from real paintings, encompassing varied artistic elements such as partial and full facial details, objects, landscapes, and background elements like the sky and trees. This selection process aimed to create a balanced representation of the artist's style rather than focusing on specific objects, ensuring that the model learned stylistic features effectively.

Generative models were fine-tuned using the DreamBooth \cite{b19} technique with Stable Diffusion to produce synthetic datasets that reflect the unique style of each artist. The model was trained with checkpoints varying from 1000 to 6000 steps, and a visual evaluation of the generated images indicated that the outputs at 6000 steps provided a more realistic and refined representation of the artists’ styles while avoiding overfitting. The instance prompt used in all experiments was ``a [TOKEN] painting", where [TOKEN] is a unique sequence of letters assigned to each fine-tuned model specific to an author.

Once training was complete, 100 synthetic images per artist were generated using the ComfyUI interface \cite{b20}. To ensure consistency across artists, the same positive and negative prompts were applied to all generations. The positive prompt, ``A full [TOKEN] painting", was chosen to encourage the generation of complete compositions rather than isolated or cropped elements. Additionally, the negative prompt explicitly included terms such as ``cropped body", ``partial figures", ``close-ups", and ``incomplete", aiming to refine the stylistic quality of the outputs and reduce unwanted framing artifacts. However, despite these efforts, the generated images still exhibited a tendency toward cropped compositions.

\begin{figure*}[!htbp]
\centering
\includegraphics[width=0.95\linewidth]{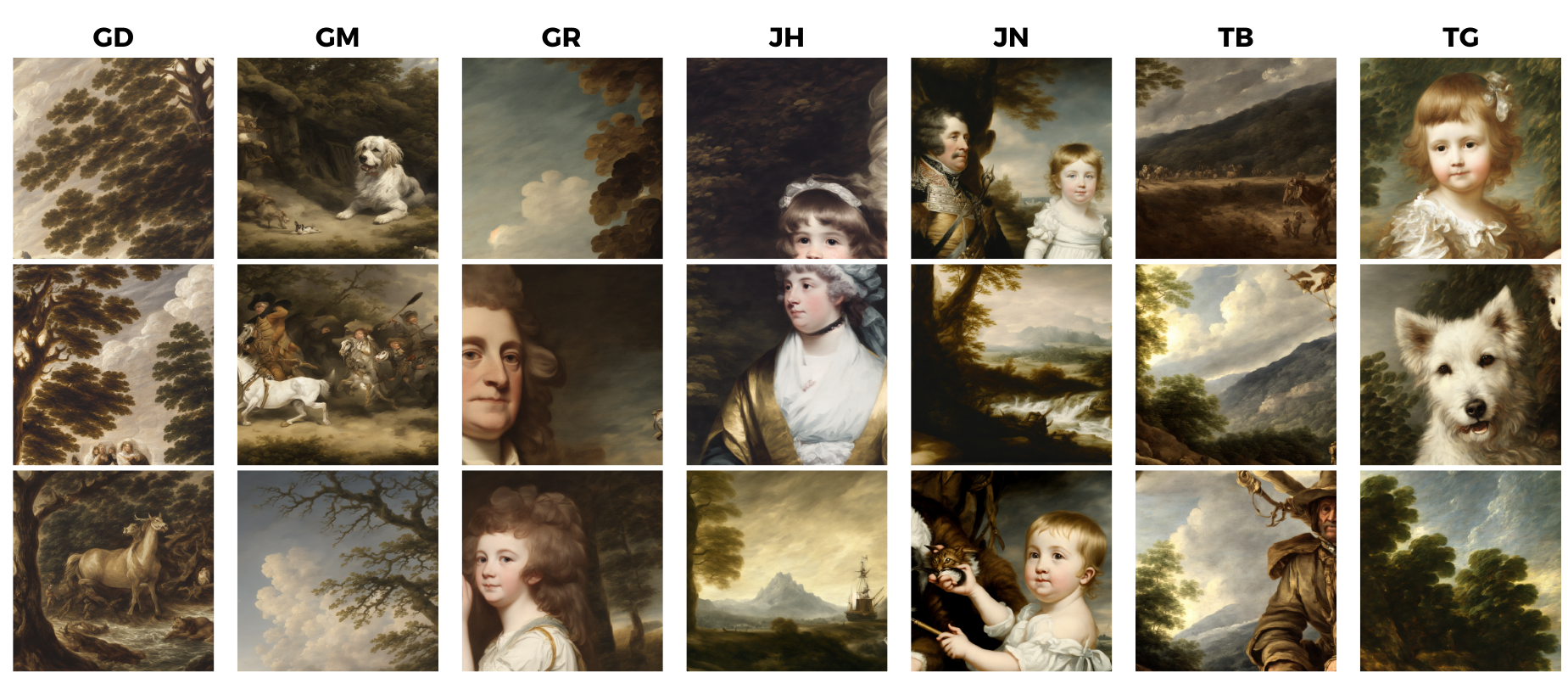}
\caption{Examples of three synthetic images generated for each artist. The images reflect stylistic variations captured by the model, showcasing differences in composition, subject matter, and brushwork. Despite efforts to mitigate generation biases, some partial figures and cropped elements remain, likely influenced by the characteristics of the training dataset.}
\label{fig:synthetic_images}
\end{figure*}

Figure \ref{fig:synthetic_images} presents examples of the images generated by our models. This result suggests that, although the prompt was designed to mitigate this effect, the bias was likely inherited from the real paintings used for fine-tuning, many of which already contained partial elements and cropped figures. Consequently, the model learned and reproduced these framing patterns, reinforcing the notion that the structure of the training data strongly influences generative outputs. Nevertheless, while the generated objects resemble the style of real paintings, they do not replicate specific figures from the original artworks, indicating that the model successfully captured stylistic elements rather than memorizing paintings.

\subsection{Training Discriminative Models}

To perform authorship attribution, we trained a single discriminative model that utilizes the combined embeddings extracted from VOLO, BEiT v2, and MaxViT. By leveraging embeddings, we transformed complex visual information into a more compact and informative representation, which enhances classification performance by focusing on high-level features rather than raw pixel data. This approach also reduces computational complexity and improves generalization, as it abstracts away from low-level noise and variations inherent in individual images. Instead of training separate models for each embedding type, we integrated all three feature representations into a unified classification model per artist, leveraging a richer and more diversified feature space.

In the Real-Only, Synthetic-Real and Hybrid experiments, a single real painting per artist was reserved for testing, ensuring that the same image was consistently used across both settings. For the Synthetic-Only experiment, the test set comprised five synthetic images per artist, specifically the last five images from each synthetic dataset, maintaining consistency in evaluation.

For classification, we trained a LightGBM model using gradient boosting decision trees. The decision to use LightGBM was driven by its efficiency in handling large datasets and its ability to capture complex relationships in the data. The training process incorporated class weight balancing to mitigate data imbalance, a learning rate of 0.05, and early stopping to prevent overfitting. Model optimization was guided by the AUC metric, with binary log loss serving as an additional stopping criterion. Once trained, the model’s performance was evaluated based on the AUC metric, providing a robust assessment of its ability to distinguish artistic styles across different experimental settings.

\section{Results and Discussion}

We evaluated five experimental settings to understand how different combinations of real and synthetic data impact model performance. The Synthetic-Only setting achieved the highest overall results, with ROC-AUC values above 0.98 and accuracy exceeding 95\% across all artists. In contrast, the Synthetic-to-Real setting showed the lowest scores, exposing a domain gap that hindered generalization, especially for TB and GM. The Real-Only setting yielded moderate performance, while the Hybrid configurations (M1 and M2) consistently improved over the baseline. Among them, Hybrid-M2 produced the most stable and robust results. Artists GD and GM performed well across all settings, while JH and TB showed greater sensitivity to training configuration.

\begin{figure}[htbp]
\centering
\includegraphics[width=0.9\linewidth]{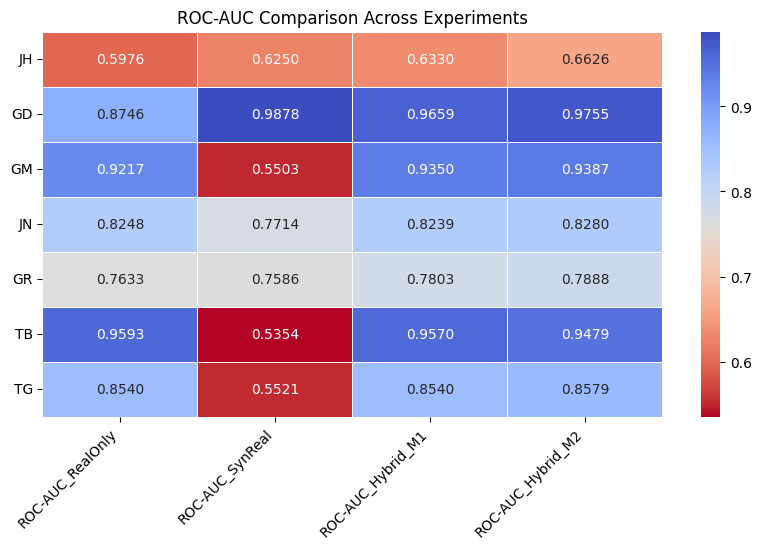}
\caption{Heatmap of ROC-AUC values for different training settings. The figure illustrates the variation in the model's ability to discriminate between classes depending on the dataset used for training and testing. Higher values indicate better discrimination between authors.}
\label{fig:auc}
\end{figure}

\begin{figure}[htbp]
\centering
\includegraphics[width=0.9\linewidth]{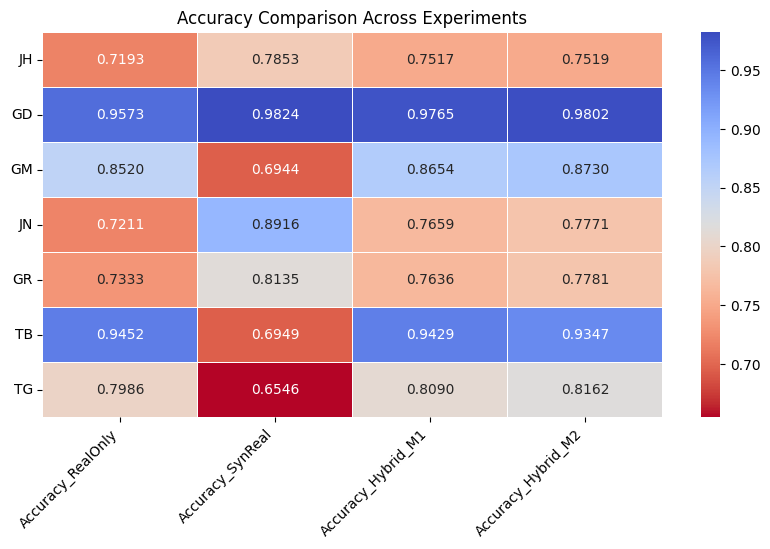}
\caption{Heatmap of Accuracy values across training settings. The figure highlights differences in overall classification performance, showing how accuracy varies depending on whether the model was trained with real, synthetic, or mixed data.}
\label{fig:acc}
\end{figure}

Figures~\ref{fig:auc} and~\ref{fig:acc} visualize ROC-AUC and accuracy results per artist and experiment, highlighting how synthetic data improves model generalization. Table~\ref{tab:performance_comparison} summarizes the best and worst cases for each configuration.

\begin{table}[htbp]
\caption{Performance comparison per experiment: best and worst artist-specific results across evaluation metrics}
\label{tab:performance_comparison}
\begin{center}
\resizebox{\linewidth}{!}{
\begin{tabular}{|c|c|c|c|c|c|}
\hline
\textbf{Experiment} & \textbf{Metric} & \makecell{\textbf{Most Accurate} \\ \textbf{Artist}} & \textbf{Best} & \makecell{\textbf{Least Accurate} \\ \textbf{Artist}} & \textbf{Worst} \\
\hline

\multirow{5}{*}{Real-Only} 
& ROC-AUC & TB & 0.9593 & JH & 0.5976 \\
& Precision & GD & 0.9644 & JH & 0.8346 \\
& Recall & GD & 0.9919 & JN & 0.7163 \\
& F1-Score & GD & 0.9780 & JH & 0.8288 \\
& Accuracy & GD & 0.9573 & JH & 0.7193 \\
\hline

\multirow{5}{*}{Synthetic-Only} 
& ROC-AUC & GM & 0.9991 & JN & 0.9833 \\
& Precision & GM & 0.9938 & JN & 0.9671 \\
& Recall & GM & 0.9938 & JN & 0.9819 \\
& F1-Score & GM & 0.9938 & JN & 0.9744 \\
& Accuracy & GM & 0.9893 & JN & 0.9558 \\
\hline

\multirow{5}{*}{Synthetic→Real} 
& ROC-AUC & GD & 0.9878 & TB & 0.5354 \\
& Precision & GD & 0.9868 & TG & 0.8417 \\
& Recall & GD & 0.9949 & TB & 0.7333 \\
& F1-Score & GD & 0.9908 & TB & 0.8121 \\
& Accuracy & GD & 0.9824 & TG & 0.6546 \\
\hline

\multirow{5}{*}{Hybrid M1} 
& ROC-AUC & GD & 0.9659 & JH & 0.6330 \\
& Precision & GD & 0.9806 & JH & 0.8520 \\
& Recall & GD & 0.9952 & JN & 0.7675 \\
& F1-Score & GD & 0.9878 & JN & 0.8609 \\
& Accuracy & GD & 0.9765 & JH & 0.7517 \\
\hline

\multirow{5}{*}{Hybrid M2} 
& ROC-AUC & GD & 0.9756 & JH & 0.6627 \\
& Precision & GD & 0.9868 & JH & 0.8604 \\
& Recall & GD & 0.9927 & JN & 0.7799 \\
& F1-Score & GD & 0.9897 & JH & 0.8475 \\
& Accuracy & GD & 0.9803 & JH & 0.7520 \\
\hline

\end{tabular}
}
\end{center}
\end{table}

These results suggest that some artists are more affected by domain shifts than others. TB, for example, showed a significant drop in AUC when transitioning from Real-Only to Synthetic-Real, indicating poor representation in the synthetic dataset. In contrast, GD maintained high performance across all settings, suggesting that his style was well preserved in synthetic generations.

\subsection{Impact of the Samples}

The M2 configuration, which doubled the sampling density from M1, led to performance improvements in most cases. For GD, the ROC-AUC increased from 0.9659 in Hybrid-M1 to 0.9756 in Hybrid-M2, and accuracy rose from 0.9765 to 0.9803. Similar gains were observed for GM and TG. For JH, who had struggled with M1 (ROC-AUC 0.6330), the score improved to 0.6627 in M2, narrowing the gap compared to Real-Only. On the other hand, TB saw a drop in ROC-AUC from 0.9570 to 0.9480, although accuracy slightly improved. Overall, these findings indicate that denser patch extraction strengthens generalization, particularly for artists with scarce training data, though its effectiveness remains limited when synthetic images fail to capture the finer stylistic traits of a painter.

\subsection{Effect of Real Data Availability}

The benefit of synthetic data varied with the number of real paintings available per artist. For those with fewer real samples, gains were substantial. For instance, GD (7 paintings) improved from 0.8746 in Real-Only to 0.9756 in Hybrid-M2 (ROC-AUC), with accuracy rising from 0.9573 to 0.9803. Similarly, JH (9 paintings) increased from 0.5976 to 0.6627 in ROC-AUC and from 0.7193 to 0.7520 in accuracy. These results highlight synthetic augmentation as an effective few-shot amplifier.

In contrast, artists with larger datasets showed more modest or inconsistent improvements. GM (22 paintings) increased slightly in ROC-AUC (0.9217 → 0.9387) and accuracy (0.8520 → 0.8730), while TG (23 paintings) showed marginal changes (0.8540 → 0.8580 ROC-AUC; 0.7986 → 0.8163 accuracy). TB (14 paintings) represents an exception: despite strong Real-Only performance (ROC-AUC 0.9593), scores dropped to 0.5354 in Synthetic-to-Real and 0.9480 in Hybrid-M2, suggesting that his style was poorly captured by synthetic generations. 

Overall, these findings suggest diminishing returns when real datasets are sufficiently large: synthetic samples add less new information and may even introduce bias or noise. Synthetic augmentation is therefore most impactful in data-scarce regimes, where it broadens stylistic coverage, balances classes, and regularizes the classifier through added diversity.

\section{Conclusion}

This work investigated the use of synthetic images to improve authorship attribution models in the context of limited and stylistically overlapping datasets. By combining real and synthetic data through a Hybrid approach, we demonstrated consistent gains in classification performance, particularly for artists with few available paintings. The results show that denser patch sampling (M2) and dataset augmentation are effective strategies for enhancing generalization. However, the performance varied across artists, indicating that the benefits of synthetic data depend on both dataset size and the quality of the generated images. A noticeable domain gap in the Synthetic-to-Real setting also revealed the limitations of current generation techniques.

These findings underscore the potential of synthetic data as a complementary resource in art authentication, while highlighting the need for adaptive sampling, improved generative models, and domain adaptation methods. Future work should explore strategies that dynamically adjust the amount and diversity of synthetic data per artist, aiming for higher fidelity and more robust cross-domain performance. Further, since all artists were assigned the same number of synthetic samples independently of their original dataset size, future work could explore adaptive sampling strategies. In summary, synthetic data appears to help by broadening stylistic coverage, balancing classes for artists with fewer real samples, and injecting diversity that regularizes the classifier.

\section*{Code and Data availability}
The code and data used for the machine-learning analyses, available for non-commercial use, have been deposited at Figshare: \url{https://doi.org/10.6084/m9.figshare.30168082}.

\vspace{12pt}

\end{document}